\documentclass[10pt,twocolumn,letterpaper]{article}

\usepackage[pagenumbers]{cvpr} 

\usepackage[accsupp]{axessibility}  
\usepackage{graphicx}
\usepackage{amsmath}
\usepackage{amssymb}
\usepackage{booktabs}
\usepackage{mathrsfs}
\usepackage{makecell}
\usepackage{comment}
\usepackage{multirow}
\usepackage{adjustbox}
\usepackage{picins}
\usepackage{wrapfig}
\usepackage{appendix}
\DeclareMathOperator*{\argmax}{arg\,max}

\usepackage[pagebackref,breaklinks,colorlinks]{hyperref}

\usepackage[capitalize]{cleveref}
\crefname{section}{Sec.}{Secs.}
\Crefname{section}{Section}{Sections}
\Crefname{table}{Table}{Tables}
\crefname{table}{Tab.}{Tabs.}
 
\usepackage[dvipsnames]{xcolor}

\newcommand{\bg}{\mathbf{g}}

\newcommand{\bp}{\mathbf{p}}
\newcommand{\bq}{\mathbf{q}}

\newcommand{\bt}{\mathbf{t}}
\newcommand{\bu}{\mathbf{u}}
\newcommand{\bv}{\mathbf{v}}

\newcommand{\bx}{\mathbf{x}}

\newcommand{\bA}{\mathbf{A}}

\newcommand{\bD}{\mathbf{D}}

\newcommand{\bF}{\mathbf{F}}
\newcommand{\bG}{\mathbf{G}}

\newcommand{\bI}{\mathbf{I}}

\newcommand{\bM}{\mathbf{M}}

\newcommand{\bP}{\mathbf{P}}
\newcommand{\bQ}{\mathbf{Q}}
\newcommand{\bR}{\mathbf{R}}

\newcommand{\bU}{\mathbf{U}}
\newcommand{\bV}{\mathbf{V}}

\newcommand{\mH}{\mathcal{H}}

\def\R{{\cal R}}

\def\U{{\cal U}}
\def\V{{\cal V}}

\def\cvprPaperID{1707} 
\def\confName{CVPR}
\def\confYear{2022}

\begin{document}

\title{APES: Articulated Part Extraction from Sprite Sheets
\vspace{-7pt}
}

\author{ 
    \hspace{-7mm}Zhan Xu$^{1,2}$
    \,
    Matthew Fisher$^2$
    \,
    Yang Zhou$^2$ 
    \,
    Deepali Aneja$^2$
    \,
    Rushikesh Dudhat$^1$
    \,
    Li Yi$^3$
    \,
    Evangelos Kalogerakis$^1$
    \\
   \hspace{-3mm}$^1$University of Massachusetts Amherst \,\,\,\,\, $^2$Adobe Research
    \,\,\,\,\,$^3$Tsinghua University
\vspace{-7pt}      
}

\maketitle

\begin{abstract}
\vspace{-7pt}  
   Rigged puppets are one of the most prevalent representations to create 2D character animations. Creating these puppets requires partitioning characters into independently moving parts. In this work, we present a method to automatically identify such articulated parts from a small set of character poses shown in a sprite sheet, which is an illustration of the character that artists often draw before puppet creation. Our method is trained to infer articulated parts, e.g. head, torso and limbs, that can be re-assembled to best reconstruct the given poses. Our results demonstrate significantly better performance than alternatives qualitatively and quantitatively.Our project page
  \url{https://zhan-xu.github.io/parts/} includes our code and data.
\end{abstract}

\vspace{-7pt}
\section{Introduction}
\label{sec:intro}
Creating rich, animated characters has traditionally been accomplished by independently drawing each frame of the character. To accelerate this process, tools have been developed to allow precisely rigged 2D characters to be easily rendered in different poses by manipulating the rig. To create these rigs, artists often start by drawing several different poses and configurations of the complete character in a sprite sheet or turnaround sheet. They then manually segment out the common parts in these sheets and stitch them together to create the final character rig, which can then be articulated to reconstruct the original character drawings\cite{liu2014skinning}. The obtained parts from different sprite sheets can also be used as assets and assembled freely to create new character rigs\footnote{https://pages.adobe.com/character/en/puppet-maker}.

Significant expertise is required to create a well-rigged 2D character, and automatic rigging methods have several unique challenges. Animated characters can have a wide range of different limbs, accessories, and viewing angles, which prevents a single template from working across all characters. Furthermore, the amount of available examples for rigged, animated characters is relatively small when compared against real datasets that can be acquired by motion capture or other techniques. This limited data is particularly challenging to work with because characters are often drawn and animated in different styles. Finally, poses shown in sprite sheets have both articulated variation and non-rigid deformation. Extracting articulated parts that express given poses requires effective analysis of the motion demonstrated in sprite sheets.

\begin{figure}
     \centering
     \includegraphics[width=0.5\textwidth]{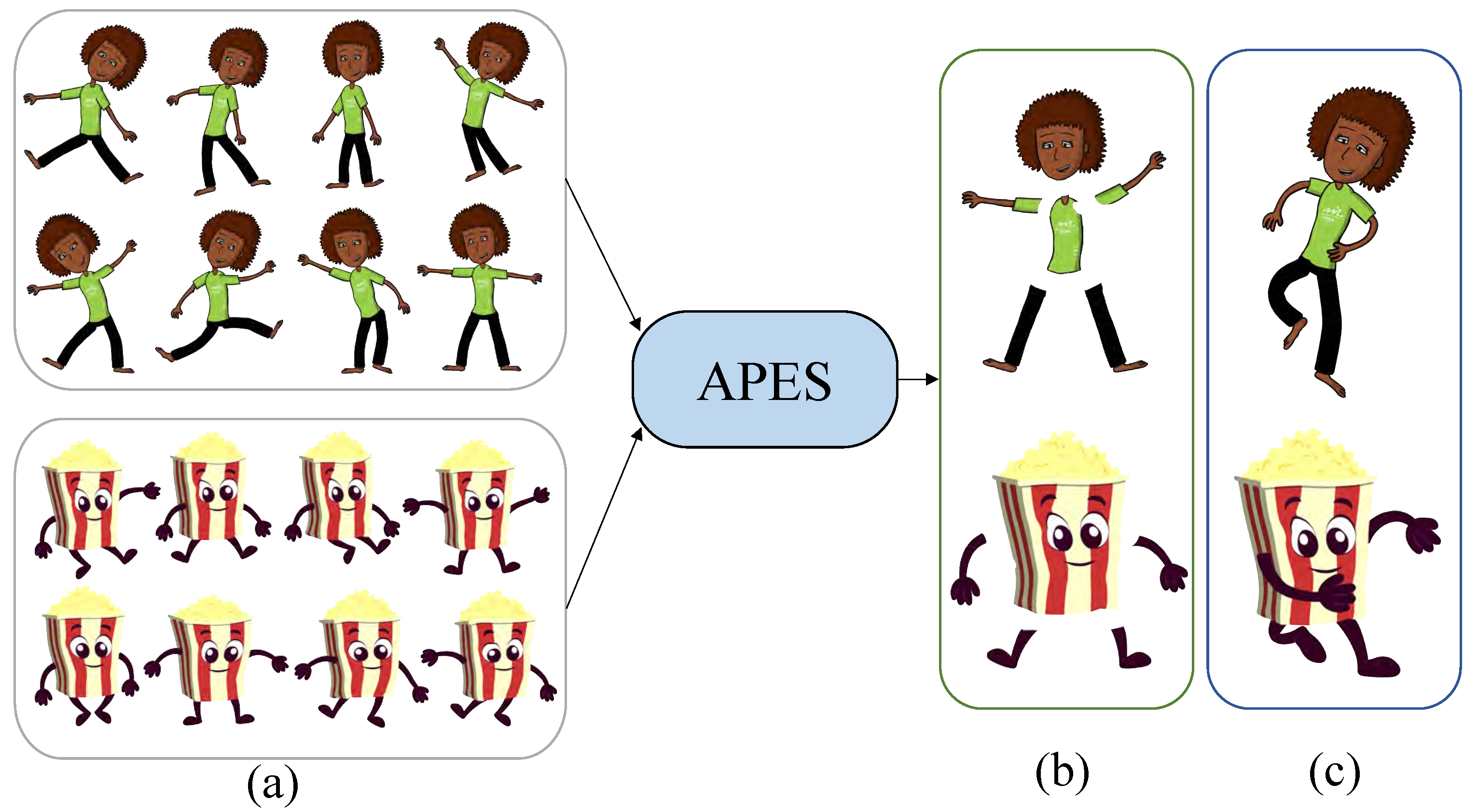}
     \vspace{-15pt}
     \caption{Given sprite sheets as input (a), APES produces articulated parts (b) that can best express poses in the sprite sheets. The obtained parts can further be warped to generate new poses (c), or manipulated freely to create new puppets.}
     \label{fig:teaser}
     \vspace{-5pt}
\end{figure}

We propose a method to automatically construct a 2D character rig from a sprite sheet containing a few examples of the character in different poses. 
Our rig is represented as a set of deformable layers~\cite{willett2017secondary}, each capturing an articulated part. We assume that all characters in the sprite sheet can be reconstructed by applying a different deformation to each puppet layer and then compositing the layers together. We start by learning a deep network that computes correspondences between all pairs of sprites. We then use these correspondences to compute possible segmentations of each sprite. Finally, we attempt to reconstruct the other sprites in the sprite sheet using the possible puppet segmentations, choosing the set with minimal overall reconstruction error.
 
We evaluate our method on several test sprite sheets. We show that our method can successfully produce articulated parts  and significantly outperforms other representative appearance and motion-based co-part segmentation works\cite{Hung_2019_CVPR,siarohin2021motion}. Our contributions are the following:

\begin{itemize}
  \vspace{-5pt}
  \item A method for analyzing a sprite sheet and creating a corresponding articulated character that can be used as a puppet for character animation.
  \vspace{-5pt}
  \item A neural architecture to predict pixel motions and cluster pixels into articulated moving parts without relying on a known character template.
  \vspace{-5pt}
  \item An optimization algorithm for selecting the character parts that can best reconstruct the given sprite poses. 
\end{itemize}

\section{Related Work}
\label{sec:related_work}
\paragraph{Rigid motion segmentation.} Several approaches \cite{tokmakov2017learning, tokmakov2019learning,tron2007benchmark,vidal2006two,yuan2007detecting,song2017embedding}  have been proposed to cluster pixels into groups following similar rigid motions.
One line of work \cite{tokmakov2017learning, tokmakov2019learning} identifies rigid groups by discovering distinct motion patterns from 2D optical flow. These methods typically work well on smooth video sequences, but cannot generalize to images with large pose changes between each other.  
They also aim at object level segmentation, and often miss articulated parts within each object. 
Other works employ 3D geometric constraints and features to infer the underlying motion of pixels for clustering~\cite{tron2007benchmark,vidal2006two,yuan2007detecting, song2017embedding,yang2021learning}. These methods also assume small motions, and  require multi-view input to perform 3D geometric inference, thus are not applicable to artistic sprites. 

\vspace{-10pt}

\paragraph{Co-part segmentation.} 
Several works focus on 
segmenting common foreground objects or parts from a set of images or video frames~\cite{vicente2011coseg,tsai2016semantic,joulin2012multi,zhang2021cyclesegnet,collins2018deep, Hung_2019_CVPR,lu2020_pami,choudhury21unsupervised}. They often utilize features from pretrained networks on ImageNet\cite{russakovsky2015imagenet}, thus are more suitable for natural images instead of non-photorealistic images, such as sprites. Most importantly, their  segmentation relies more on appearance and semantic consistency rather than part motion. As a result, they miss articulated parts, even when trained on our datasets, 
as shown in our experiments in the case of SCOPS \cite{Hung_2019_CVPR}.

Other co-part segmentation works rely on motion to better extract articulated parts. 
Early methods use keypoint tracking and various strategies for trajectory recovery and modeling \cite{Wang93,yan2008factorization,Ochs11,del2016discovering, chang2017highly}. However, they are often hand-tuned and prone to noisy tracking and large pose deformations. More recently, deep learning methods have shown promising results for motion-based co-segmentation
\cite{psd,sabour2021unsupervised,siarohin2021motion}. However, they heavily rely on well-predicted optical flow. When input images have distinct and large pose changes, optical flow becomes unreliable. They are also more suitable for natural images of objects from a single category. When trained on sprite sheets with varying articulation structure, they produce unsatisfactory results, as shown in our experiments for the recent approach of \cite{siarohin2021motion}.

\vspace{-10pt}

\noindent
\paragraph{3D mobility segmentation.} Mobility-based segmentation for 3D point clouds has also been investigated in recent works
\cite{li2007projective,li2016mobility,xu20193d,Wang_2019_CVPR}. Yi \emph{et al.} \cite{yi2018deeppart} predicted point cloud segmentation from a pair of instances under different object articulation states. Hayden et al. \cite{hayden2020nonparametric} proposed an unsupervised  part model to infer parts in a 3D motion sequence. MultiBodySync \cite{huang2021multibodysync} achieved consistent correspondence and segmentation from multiple articulation states of the same object by spectral synchronization. All these approaches are designed for 3D point clouds or meshes. Although we are inspired by these approaches to handle large pose variations, our method incorporates several adaptations for processing 2D sprites, including a convolutional correspondence module for pixel correspondence, a neural voting strategy to handle efficient clustering of rigid motions in superpixel space, and an optimization strategy to find common parts leading to the best reconstruction of sprites.
\vspace{-13pt}
\noindent
\paragraph{Puppet rigging and deformation.} Prior works on puppet deformation \cite{Poursaeed_2020_WACV,hinz2021charactergan} assumes that the parts and their hierarchy are given i.e., the articulated parts have been specified by artists. Our approach is complementary to these methods, aiming to automate part extraction useful in their input. Recently, Xu \etal \cite{RigNet} proposed a neural network to infer a hierarchical rig for articulated characters. However, it relies only on the 3D geometry of the model, and does not take into account motion cues, as we do.

\vspace{-5pt}
\section{Method}
\label{sec:method}

 \begin{figure*}[t]
    \centering
    \includegraphics[width=1.0\linewidth]{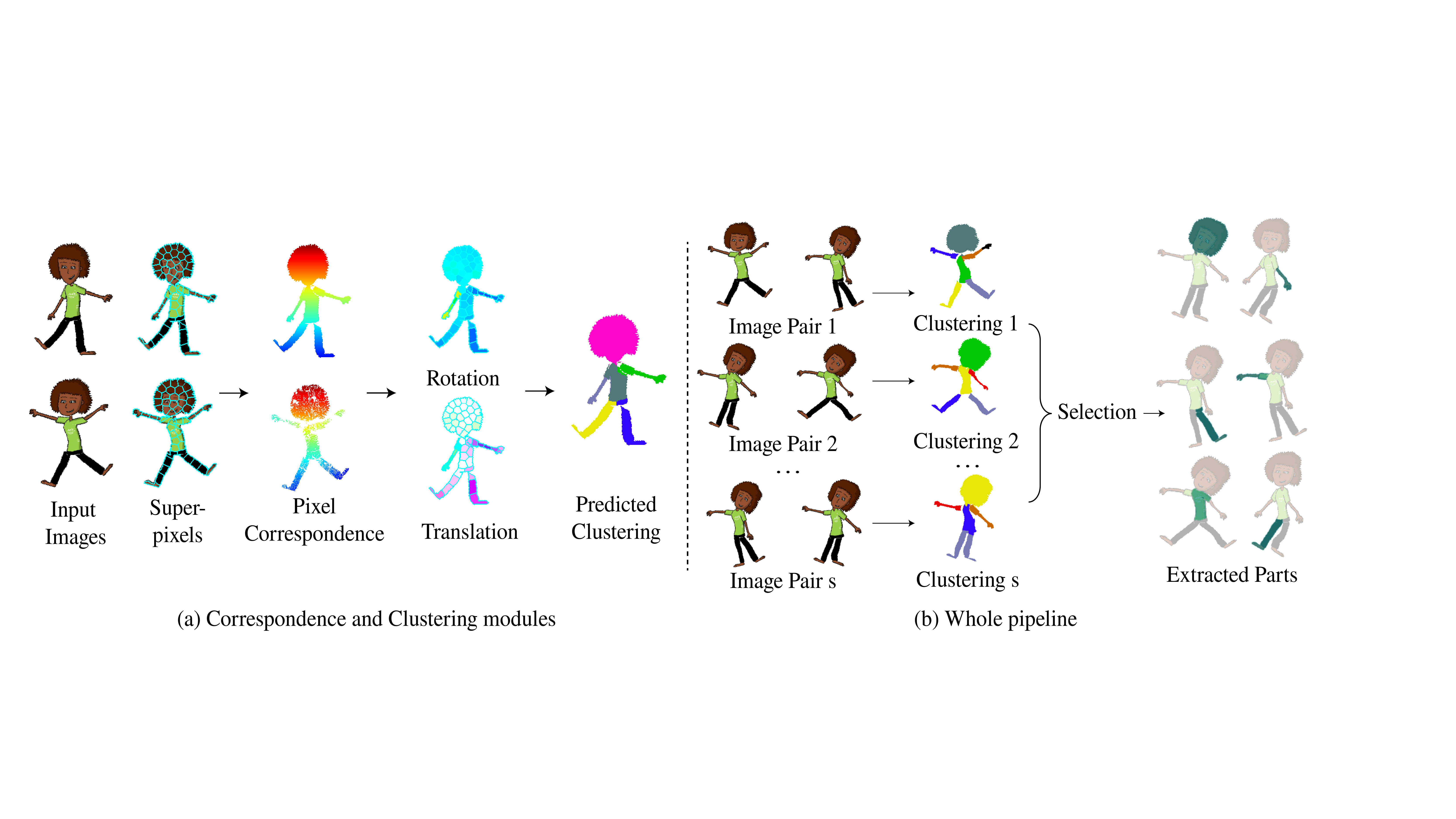}
    \vspace{-8mm}
    \caption{Pipeline overview. (a) Given any pair of images, the correspondence module (Sec.\ref{corr}) predicts candidate pixel correspondence between them. The clustering module (Sec.\ref{seg}) then groups superpixels with similar rigid transformation together. (b) After applying the correspondence and clustering modules to all pairs of images, we collected parts scattered across all poses. We select a subset through optimization that can best reconstruct the given poses while having minimal overlap between them.}
    \label{fig:overview}
    \vspace{-5mm}
\end{figure*}

The goal of our method is to infer the articulated parts of a 2D cartoon character given only a few poses drawn by an artist under different articulations. The number of poses can vary for each character, e.g. $6$ to $10$ in our datasets. The input are $P$ sprite RGB raster images $\bI_i$ and their accompanying foreground binary masks $\bM_i, i=[1,...,P]$, where $P$ is the total number of poses. The output is a set of articulated body parts that artists can subsequently animate based on standard part rigging methods and software \cite{borosan2012rigmesh,liu2014skinning,adobe_ch} (see Fig.~\ref{fig:teaser} for examples).

The pipeline of our method is shown in Fig.~\ref{fig:overview}. First, given any pair of images (poses) from the inputs, the first module of our method, i.e \emph{correspondence} module, (Sec. \ref{corr}) infers the pixel correspondences which capture the candidate motions of pixels between images.
These correspondences are then processed through the \emph{clustering} module (Sec. \ref{seg}), which attempts to find pixels with similar motion patterns and groups them into a set of candidate articulated parts as the output. This modular architecture has the advantage of \textit{disentangling} motion from appearance and using only motion patterns for clustering.
Finally, we gather candidate parts from all pairs and select a final set of parts to represent the target puppet (Sec. \ref{select}). The selected parts are required to have minimum overlap with each other and also reconstruct all poses with as-rigid-as-possible deformations.

The correspondence and clustering module have the form of neural networks that we both train jointly in a supervised manner (Sec. \ref{sec:training}) based on publicly available puppet datasets (Sec. \ref{subsec:datasets}). We observe that networks can still generalize to real, artist-made cartoon characters and poses. The part selection solves a parameter-free optimization problem which does not require training.

\subsection{Correspondence module}
\label{corr}

Given a pair of images $\bI_s, \bI_t$
and corresponding
binary foreground masks $\bM_s, \bM_t$ from the input set, the correspondence module predicts the candidate motion mapping of foreground pixels between two images. 
To achieve this, the module first concatenates each input image and mask, then transforms them into a feature map $\bF_s \in \mathbb{R}^{H \times W \times 64}$  using a convnet. 
The network follows a U-Net architecture \cite{ronneberger2015u} and consists of ten convolutional layers in its encoder and another ten layers in its decoder. The convolutional layers in the encoder implement \textit{gated convolution} \cite{yu2018free}, whose gating mechanism prevents background pixels indicated by the masks 
from influencing the foreground pixel correspondences. 
The feature vector of each pixel is normalized according to its $L_2$ norm such that it is unit length (i.e., it lies on the unit hypersphere \cite{wang2020hypersphere}).

Next, given each foreground pixel $\bx=(x, y)$ in the source image $\bI_s$, its corresponding pixel $\bx'=(x',y')$ in the target image $\bI_t$ is found as the
pixel with the most similar feature vector in terms of cosine similarity: 
\vspace{-3pt}
\begin{equation}
\label{eq:corr}
\bx' = \argmax_{ \bu \in \bI_t , \bM_t(\bu)=1}
\big( \bF_s(\bx) \cdot \bF_t(\bu) \big)
\end{equation}

\vspace{-3pt}
We experimented with alternatives to extract correspondences such as RAFT~\cite{teed2020raft} and COTR \cite{jiang2021cotr} trained on the same dataset as ours. Both resulted in worse results (see our experiments section for comparisons and discussion).

\subsection{Clustering module}
\label{seg}
Given the pixel correspondences between the source and the target image $\bI_s, \bI_t$, our clustering module aims to discover character articulated parts by grouping pixels with similar motion transformations.
Since it is not possible to estimate the transformation, i.e. 2D rotation and translation from a single pixel, we instead gather votes for transformations from pairs of corresponding points $(\bx_1, \bx_1')$ and $(\bx_2, \bx_2')$, where $\bx_1, \bx_2$ are source pixels, and $\bx_1', \bx_2'$ are their correspondences in the target image. Then we cluster these votes to discover the dominant rigid motion transformations and associated parts, similarly to Hough voting \cite{ballard1981generalizing}. 

\vspace{-15pt}
\paragraph{Voting pairs.} Gathering votes from correspondences of all possible pixel pairs $\bx_1, \bx_2$ would be computationally expensive even for moderate image resolutions. In addition, distant pixels often belong to different parts, thus, their votes would tend to be irrelevant. To accelerate computations, we apply the superpixel segmentation method SLIC\cite{slic} to our input images and assume all pixels within a superpixel share the same motion transformation.
\vspace{-15pt}

\paragraph{Rotation extraction.} To extract the rotation from pairs of correspondences, one popular method is to use the orthogonal Procrustes analysis \cite{horn1987closed}. However, through our experiments, this approach turned out not to be robust -- even slightly noisy correspondences can significantly distort the votes. Instead, we follow a convnet approach that learns to estimate the transformations from approximate correspondences. The input to our network is a map storing the voting pairs. Specifically, for each source pixel $\bx_1$, we store the 2D vector $\bx_1-\bx_c$ representing its relative position with respect to its superpixel centroid $\bx_c$, and also the corresponding 2D vector $\bx_1'-\bx_c'$. This results in a $H \times W \times 4$ input voting map. Pixels without any correspondences are indicated by an additional binary mask.

The voting and mask maps are processed through a U-Net backbone and gated convolutions similar to the convnet of our correspondence module.  The output is a $H \times W \times 64 $ feature map representing motion features per pixel in the source image. We then apply average pooling spatially over each superpixel area to acquire motion features $\mathbb{R}^{K_s \times 64}$ for all $K_s$ superpixels. Finally, an MLP layer is applied to map the features to $\mathbb{R}^{K_s \times2}$ space, representing (residual) sine and cosine of the rotation angles for $K_s$ superpixels.

\vspace{-12pt}

\paragraph{Translation extraction.} Directly predicting both translation and rotation is possible, however we found it is more accurate to predict the rotation first, then update the motion features based on the rotation, and finally predict the translation (see also our ablation). 
In this manner, we discourage the network to express any small rotations merely as translations. The translation prediction network shares the same architecture of the rotation extraction network.

\vspace{-12pt}

\paragraph{Clustering.} Given extracted rotations and translations for superpixels, we  proceed with characterizing their motion similarity, or in other words \textit{affinity}. 
This affinity is computed based on motion residuals inspired by \cite{huang2021multibodysync}.
We apply the estimated rotation and translation of each superpixel to transform all other superpixels, and compute the position difference between the transformed superpixels and their corresponding superpixels. Specifically, given a super-pixel $\bp_i$ with extracted rotation matrix $\bR_s[i]$ and translation $\bt_s[i]$, the motion residual for the super-pixel $\bp_j$ is computed as: 
\begin{equation}
    \bD_s(i,j)=\frac{\sum_{\bx \in \bp_j} (\bR_s[i] \cdot \bx + \bt_s[i] - \bx')}{|\bp_j|}  
\label{eq:mat_d}
\end{equation}
where $|\bp_j|$ is  the number of pixels in the superpixel $\bp_j$.
The motion residual matrix
$\bD_s \in \mathbb{R} ^ {K_s \times K_s \times 2}$ is processed through more MLP layers to compute the superpixel affinity matrix  $\bA_s \in \mathbb{R} ^ {K_s \times K_s}$. More  details on the architecture can be found in the supplementary. 

Given the predicted affinity matrix $\bA_s$, the grouping is achieved by using spectral clustering~\cite{ng2002spectral}. Here  we follow the differential clustering approach \cite{arrigoni2019motion, huang2021multibodysync}, which results in matrix $\bG_s \in \mathbb{R}^{K_s \times C_s}$ representing a soft membership of superpixels to $C_s$ clusters.
We follow \cite{huang2021multibodysync} to set the number of clusters based on the number of eigenvalues extracted from spectral clustering larger than a threshold. Here we set the threshold as $1\%$ of the sum of the first 10 eigenvalues. By converting the soft membership to a hard one, the resulting clusters reveal articulated parts for the source pose based on its paired target pose.

\subsection{Part Selection}
\label{select}

By passing each pair of poses $\bI_s, \bI_t$ through  our correspondence and segmentation modules, we obtain a set of parts for the source pose $\bI_s$. Processing all pairs of poses yields a ``soup'' of candidate parts $\bQ=\{\bq_1, \bq_2,..., \bq_C\}$ scattered across all poses, where $C = \sum_s C_s$ is their total number. Obviously many of these parts are redundant e.g., the same arm extracted under different poses. Our part selection procedure  selects a compact set of parts that (a) can reconstruct all poses with minimal error, and also (b) have minimum overlap with each other. 
To reconstruct poses, one possibility is to use rigid transformations of candidate parts. Despite the fact that rigidity was used to approximately model the motion of parts in the previous section,  not all the sprite sheet characters are fully rigidly deformed. There are often small non-rigid deformations within each part and around their boundaries. Thus we resort to as-rigid-as-possible (ARAP) deformation \cite{sorkine2007rigid} for reconstructing poses using the selected parts more faithfully.

To satisfy the above criteria, we formulate a \emph{``set cover''} optimization problem where the smallest sub-collection
of ``sets'' (i.e., parts in our case) covers a universe $\bP=\{\bp_i\}$ of ``elements'' (i.e., all the superpixels across all poses). Specifically, by introducing a binary variable $z_c$ indicating whether a part $\bq_c$ belongs to the optimal set (the ``set cover'') or not, we formulate the following optimization problem:
\vspace{-5mm}
\begin{align}
&\min \sum_{\bq_c \in \bQ} z_c \nonumber \\
& s.t. \sum_{\bq_c \colon \bp_i \in \bq_c} z_c \ge 1 \,\,\,for \,\,all\,\, \bp_i\in \mathcal \bP 
\label{ilp}
\end{align}

\vspace{-3pt}
We solve the above Integer Linear Programming (ILP) problem through relaxation. This yields a continuous linear programming problem solved using the interior point method \cite{freund2004primal}. We finally apply
the randomized-rounding algorithm \cite{raghavan1987randomized} to convert the continuous result to our desired binary predictions. 
The randomized-routing can give us multiple possible solutions. We measure their quality by deforming the selected parts in each solution to best reconstruct all the given poses. The deformation is based on ARAP~\cite{sorkine2007rigid}. We choose the best solution with the minimal reconstruction error (see details in the supplementary).

\vspace{-3pt}
\section{Training}
\label{sec:training}

\vspace{-3pt}
The correspondence and clustering modules are involved in our training procedure.
\vspace{-12pt}
\paragraph{Correspondence module supervision.} 
We train the correspondence module through a contrastive learning approach using supervision of pair-wise pixel correspondences. Specifically, given a pair of input images $\bI_s, \bI_t$, we minimized a correspondence loss \cite{oord2018representation,Neverova2020} that encourages the representation of ground-truth corresponding pixel pairs $(\bx,\bx')$ to be more similar than non-corresponding ones:
\begin{equation}
    L^{(corr)}_{\bx , \bx'} =
    -\log\frac{\exp{ \big( \bF_s(\bx) \cdot \bF_t(\bx') /\tau \big) }}
    {\sum\limits_{\bu \in \bU_t}
    \exp{ \big( \bF_s(\bx) \cdot \bF_t(\bu)/\tau \big)}}
\label{eq:infonce}
\end{equation}
where $\bU_t$ is a predefined number of pixels we randomly sample from the foreground region of the image $\bI_t$ indicated by its mask $\bM_t$. We set this number to $1024$ in our experiments. The temperature $\tau$ is used to scale the cosine similarities. It is initially set as 0.07, and is learned simultaneously as we train the correspondence module~\cite{radford2021learning}. The total correspondence loss $L_c$ is averaged over all training corresponding pixel pairs.

 During training, we alternatively replace the  argmax of Eq.~\ref{eq:corr} with a soft version 
 to preserve differentiability and enable backpropagation of losses from the clustering model. Specifically, we replace it with 
 the weighted average of the top-$\kappa$ closest target image foreground pixels to each source image pixel ($\kappa=3$ in our implementation):
\begin{equation}
\bx' =  
\frac
{
\sum_{\bu_{\kappa} \in \U(\bx) }
\exp{ \big( \bF_s(\bx) \cdot \bF_t( \bu_{\kappa} ) /\tau\big)} \cdot \bu_{\kappa}
}
{
\sum_{\bu_{\kappa} \in \U(\bx) }
\exp{\big( \bF_s(\bx) \cdot \bF_t(  \bu_{\kappa})/\tau\big)}
}
\end{equation}
where $\U(\bx)$ represent the top-$\kappa$ most similar target pose pixels to $\bx$ using cosine similarity. The closest pixels are updated after each forward pass through our network.

\begin{figure*}[t!]
    \centering
    \includegraphics[width=1.0\linewidth]{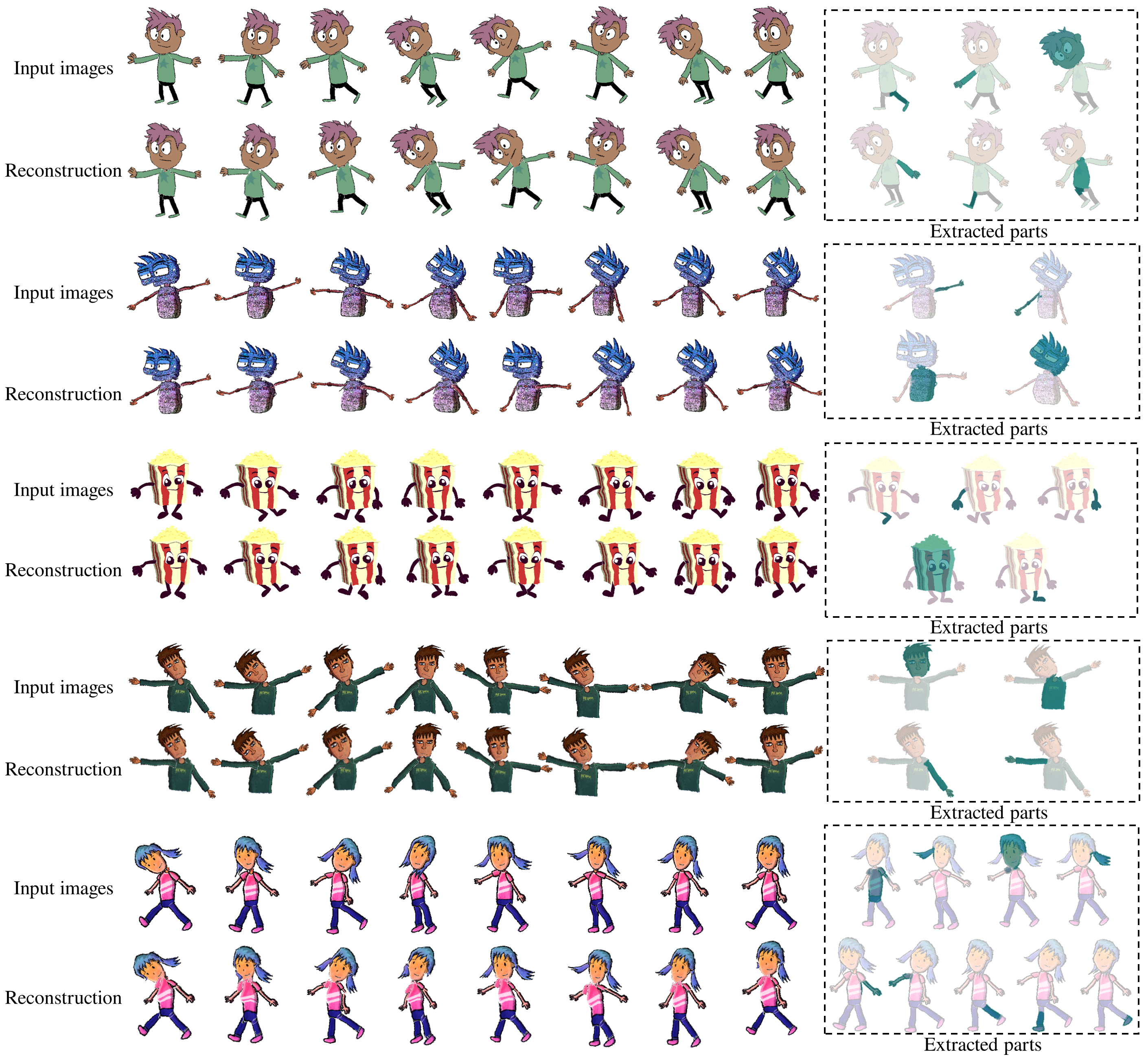}
    \vspace{-20pt}
    \caption{ \emph{Left:} The top row of each OkaySamurai test puppet shows the input poses of the sprite sheet. The bottom row shows our reconstructed poses. \emph{Right (box):} The predicted articulated parts from APES.}
    \label{fig:rec-comp-os}
    \vspace{-1em}
\end{figure*}

\vspace{-12pt}
\paragraph{Clustering module supervision.}

We train the clustering module with the binary cross-entropy (BCE) loss over the supervision of ground-truth affinity matrix $\bA_s^{gt}(i,j)$.
\begin{equation}
    L_s^{(aff)} = BCE(\bA_s, \bA_s^{gt})
\label{eq:affinity}
\end{equation}

Similarly to \cite{huang2021multibodysync}, we introduce an additional loss on the motion residual matrix $\bD_s$ to encourage consistent rigid transformation predictions, i.e. $\bR_s, \bt_s$ in Eq.~\ref{eq:mat_d}, across superpixels of the same part:
\begin{equation}
    L_s^{(motion)} = \frac{
    \sum\limits_{i,j}
    [ \bA_s^{gt}(i,j)=1 ] \cdot \|\bD_s(i,j)\|^2}
    {\sum\limits_{i,j}
    [ \bA^{gt}(i,j)=1 ] }
\label{eq:motion}
\end{equation}
where $[\cdot]$ is an indicator function. 

Finally, we adopt the soft IoU loss \cite{krahenbuhl2013parameter} to push the clustering memberships of superpixels in matrix $\bG_s$ to be as similar as possible to the ground-truth ones $\bG_s^{gt}$.
\begin{equation}
\label{eq:iouloss}
    L_s^{(clust)} = 
    \sum_{c=1}^{ C_s^{(gt)} }
    \frac{\left<\bg_c, \bg^{gt}_{\mH(c)}\right>}
    {\|\bg_c\|_1+\|\bg^{gt}_{\mH(c)}\|_1-\left<\bg_c, \bg^{gt}_{\mH(c)}\right>} 
\end{equation}
where $\bg_c$ and $\bg_c^{gt}$  represent the column of the $\bG_s$ and $\bG_s^{gt}$ respectively. $C_s^{gt}$ is the total number of parts in the ground-truth. $\mH(c)$ represents the matched column index $c$ of predicted cluster to the ground-truth cluster based on Hungarian matching \cite{kuhn1955hungarian}. 

We note that the ILP solution does not participate in our network training implementation. End-to-end training would require methods for differentiating ILPs \cite{MIPaaL,IntOpt}, yet these would make  training computationally too expensive.

\vspace{-12pt}
\paragraph{Implementation Details.}
The correspondence and clustering modules are trained using the Adam optimizer using the sum of all the above losses.
We refer readers to the  supplemental for more details, and also to our project page for source code (the link is included in our abstract).

\section{Experiments}
\label{sec:experiments}
In this section, we discuss our dataset and results. We also show qualitative and quantitative comparisons.

\subsection{Datasets}
\label{subsec:datasets}
To provide supervision to our neural modules, we make use of two publicly available datasets.

\vspace{-12pt}
\paragraph{OkaySamurai dataset.} First, we use the publicly available puppets from the OkaySamurai website\footnote{\url{https://www.okaysamurai.com/puppets/}}.
The dataset consists of $57$ artist-created and rigged characters, with varying numbers of articulated parts and spanning different categories such as full or half body humanoids, dolls, robots, often having accessories such as clothes and hand-held objects. The advantage of this dataset is that the rigged characters are already segmented into parts, which can be used to train our neural modules and allow numerical evaluation. We split the data such that $30$ puppets are used for training, $7$ for hold-out validation, and $20$ for testing. For each training and validation puppet, we generate 200 random poses and sample 100 pose pairs to train the correspondence and clustering modules. The different poses are created by specifying random angles in a range $[-0.3\pi, 0.3\pi]$ to their skeletal joints. We also apply small, additional non-rigid deformations on each body part to improve the pose diversity (see  supplementary for details).

\vspace{-12pt}
\paragraph{Creative Flow+ dataset.} Despite augmentation for poses, the number of training puppets in the OkaySamurai remains limited. Since our correspondence module is appearance-sensitive, we can pre-train it separately on larger datasets with ground-truth correspondences. One such example is the recent Creative Flow+ dataset \cite{shugrina2019creative}. The dataset contains 2D artistic, cartoon-like renderings of animation sequences along with ground truth pixel-wise correspondences. The dataset does not contain segmentation of articulated parts, yet, it is still a useful source to pretrain our correspondence module.  
The animation sequences are generated from various 3D\ meshes. We removed the ones having no articulated pose structure, e.g., the ones generated from ShapeNet models.
Since we are also interested in training on pose pairs with large motion variations, we sample poses at least 30 frames in between.
In total, we pick $8058$ pairs of CreativeFlow+ cartoon renderings for training, $1165$ pairs for validation, and $1078$ pairs for evaluating our correspondences against alternatives.

\vspace{-12pt}
\paragraph{SPRITES dataset.} We use one more dataset to evaluate how well our method generalizes to other data not involved in our training. 
We obtained $10$  sprite sheets manually created by artists\footnote{we obtained permission to publish them}.
We refer to this dataset as ``SPRITES''. For each sprite sheet, we gathered $6$-$10$ poses  of the character, all artist-drawn. The characters of this dataset  are not rigged, nor segmented into parts, thus we use this dataset for qualitative evaluation.

\vspace{-12pt}
\paragraph{Training strategy.} 

We first pre-train our correspondence module using the InfoNCE loss of Eq.~\ref{eq:infonce} on the CreativeFlow+ dataset. Starting from the pre-trained correspondence module, we then train both neural modules on the training split of the OkaySamurai dataset. This strategy offered the best performance. We also found helpful to apply color jittering augmentation to each training pair.

\begin{figure*}[t!]
     \centering
     \includegraphics[width=1.0\textwidth]{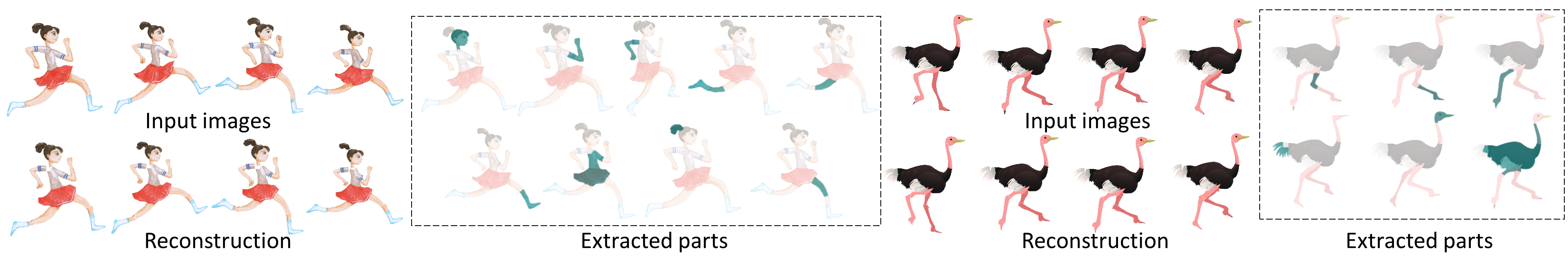}
     \vspace{-20pt}
     \caption{Part extraction from sprite sheets created by artists (``SPRITES'' dataset). In the box, we show the predicted articulated parts. }
     \vspace{-10pt}
     \label{fig:figure_real_data}
\end{figure*}

\vspace{-12pt}
\paragraph{Evaluation metrics.}

The test split from CreativeFlow+ dataset can be used for evaluating correspondence accuracy. We use the  end-point-error (\textbf{EPE}) as our evaluation protocol which measures the average distance between the predicted and the ground-truth corresponding pixels.

The test split of OkaySamurai is used to evaluate part extraction. For each testing puppet, we generate $10$ different poses to get $200$ test poses. We process all possible pairs ($45$ pairs per puppet) through our trained correspondence, clustering modules and part selection procedure to output selected articulated parts for each puppet. We also deform the selected parts by ARAP to best reconstruct each input pose (see Sec.~\ref{select}). 
For evaluating the output parts, we first perform Hungarian matching between ground-truth and reconstructed parts based on Intersection over Union (IoU), with $1-IoU$ is used as cost. The resulting average \textbf{part IoU} is used as our main evaluation metric. As additional evaluation metrics, we also use the difference between the  reconstructed and ground-truth poses by \textbf{MSE}, \textbf{PNSR} and \textbf{LPIPS}~\cite{zhang2018perceptual}. High reconstruction error indicates implausible parts used in deformation.

\subsection{Articulated Parts Selection from Sprite Sheets}

Figure \ref{fig:rec-comp-os} shows our articulated part extraction results from our method for characteristic sprite sheets from the OkaySamurai dataset. We also include reconstruction results based on the deformation procedure described in Section \ref{select} on the second row of each example. Our method successfully recovers articulated parts in most cases, although boundaries of parts are not always accurate (e.g., see shoulders and hips in the last example). 
Figure \ref{fig:figure_real_data} shows results from the SPRITES dataset. Our method is able to detect intuitive articulated parts in these artist-drawn poses, although regions near part boundaries (e.g., legs, tail of bird) are slightly grouped off. 

Our supplementary material includes additional qualitative results from the CreativeFlow+ dataset. In addition, the supplementary video shows applications of our method to automatic puppet creation and automatic synthesis of animation skeletons based on our identified parts.

\subsection{Comparisons }

\setlength{\columnsep}{10pt}%
\begin{wraptable}{r}{0.45\columnwidth}
  \begin{tabular}{@{}cc@{}}
    \toprule
    Method & IoU \\
    \midrule
    SCOPS\cite{Hung_2019_CVPR} & 27.4\% \\
    SCOPS-s (sc) & 33.1\% \\
    SCOPS-s (nosc) & 35.8\% \\
    MoCoSeg\cite{siarohin2021motion} & 26.0\% \\
    MoCoSeg-s & 32.3\% \\
    APES & \textbf{71.0\%} \\
    \bottomrule
  \end{tabular}
  \vspace{-5pt}
  \caption{Results in the OkaySamurai test set.  }
  \label{tab:comp_os}
  \vspace{-5pt}
\end{wraptable}

\paragraph{Articulated part extraction.} Our method (APES) is the first to deal with articulated part extraction from sprite sheets. There are no prior methods that have been applied to this problem. Yet, one important question is whether methods that have been developed for  part co-segmentation in photorealistic images can be applied to our problem. One may argue that appearance cues might be enough to detect the common parts across different poses of a character. To test this hypothesis, we perform comparisons with  
 \textbf{SCOPS} \cite{Hung_2019_CVPR}, a state-of-the-art co-part segmentation method. The method is self-supervised, and the self-supervision is applied to real-world images. We train SCOPS on the same training sources with our method (CreativeFlow+ and OkaySamurai), and  we also add 
 supervisory signal using our clustering loss. We call this supervised variant as \textbf{SCOPS-s}. We note that SCOPS
  does not make use of optical flow or external correspondences, thus APES still uses more supervision than SCOPS-s. Nevertheless, we consider useful to show this comparison, since SCOPS is a characteristic example of a method that does not consider motion cues.
 We also note that SCOPS uses a semantic consistency loss that makes segmentations more consistent across objects of the same category. We tested SCOPS-s
 with and without this loss; we refer to these variants as  \textbf{SCOPS-s (sc)}  and 
\mbox{\textbf{SCOPS-s (nosc)}}. We exhaustively tested the loss weights to find the best configuration, and select the best number of output parts ($12$ parts). We note that the output segmentation map from SCOPS includes a background region -- we ignore it in our evaluation. The resulting part regions can be evaluated with the same metrics, averaged over all poses of the test puppets of OkaySamurai. We finally note that SCOPS\ cannot perform reconstruction, thus, we report only segmentation performance in the OkaySamurai test dataset. 

Table \ref{tab:comp_os}
 presents the average part IoU for the OkaySamurai test set. Note that we also include the performance of the original self-supervised SCOPS\ approach just for reference.
All SCOPS variants have low performance e.g., APES' IoU in OkaySamurai is $71\%$, twice as high compared to the best SCOPS\ variant ($35\%$). 
The results indicate that appearance-based co-part segmentation is not effective at 
extracting articulated parts in sprite sheets.

An alternative co-segmentation approach is the one by Siarohin \emph{et al} ~\cite{siarohin2021motion}, which relies on motion cues. We denote it as \textbf{MoCoSeg}. Like SCOPS, it is self-supervised and trained on videos. Similarly, to make a fair comparison, we re-train this method on the same training sources as ours, and also add supervisory signal using both our clustering loss and our correspondence loss on its flow output. We call this supervised variant as \textbf{MoCoSeg-s}. We tuned their loss weights and output number of parts to achieve best performance in the validation split. 

Table~\ref{tab:comp_os} shows the performance of MoCoSeg-s and MoCoSeg, both of which are low. We suspect that the motion inferred by MoCoSeg is correlated to their part segmentation, which is more suitable for objects with consistent articulation structure. This indicates that such motion-based segmentation methods are not appropriate for our setting.  

\begin{figure}[t!]
    \centering
     \includegraphics[width=0.5\textwidth]{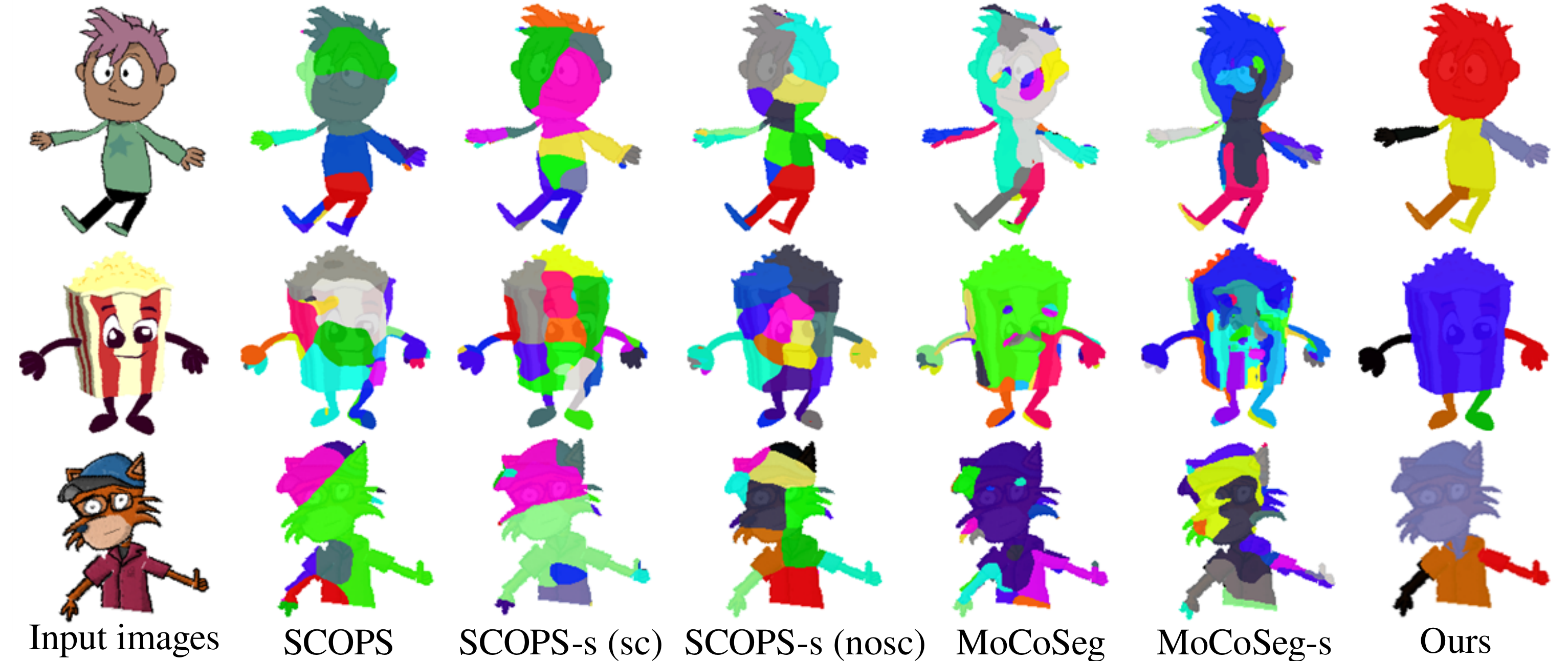}
     \vspace{-20pt}
    \caption{Identified parts from different methods on characteristic poses from the OkaySamurai dataset. Note that different color indicates different parts (colors have no semantic correspondence). APES identifies articulated parts much more successfully.}
    \label{fig:seg-comp-os}
    \vspace{-8pt}
\end{figure}

Figure \ref{fig:seg-comp-os} shows characteristic outputs from the above SCOPS\ and MoCoSeg variants and APES. Our method can  infer articulated parts from the input poses much more accurately compared to competing methods, aligning better with the underlying articulated motion.

\begin{table}
  \centering
  \begin{tabular}{cccc}
    \toprule
    Method & RAFT\cite{teed2020raft} & COTR\cite{jiang2021cotr} & APES\\
    \midrule
    EPE \small{(end-point-error)}& 28.07 & 31.93 & \textbf{22.90} \\
    \bottomrule
  \end{tabular}
  \vspace{-7pt}
  \caption{Quantitative results on the Creative Flow test split based on the EPE\ metric. Our method achieves the lowest EPE.}
  \label{tab:comp_cf}
  \vspace{-12pt}
\end{table}

\vspace{-12pt}
\paragraph{Correspondences.} We also evaluate our correspondence module against alternatives. First we compare our method with an optical flow method \textbf{RAFT} \cite{teed2020raft}. For RAFT, we remove their online-generated masks to allow longer-range optical flows and incorporate foreground masks in the correlation operation so that only foreground pixels have positive correlation. In addition, we shifted the predicted corresponding pixels to their nearest foreground pixels.
We note that we fine-tuned  RAFT on the same training datasets as ours (CreativeFlow+ and OkaySamurai training splits). This worked better compared to training it from scratch on our datasets, or
using its pre-trained model without fine-tuning.
Table~\ref{tab:comp_cf} shows  quantitative results on our test split of CreativeFlow+. Our correspondence module produces much more accurate correspondences compared to RAFT. 

We also compare to a pixel correspondence method based on transformers, called \textbf{COTR} \cite{jiang2021cotr}. We fine-tuned COTR on the same training splits as ours, and shifted the predicted corresponding pixels to their nearest foreground pixels. Still, COTR's results are inferior to APES. For visualization of correspondence results from our method and others, please see our supplementary material.

\subsection{Ablation Study}

\tabcolsep=0.10cm
\begin{table}
  \centering
  \begin{tabular}{cccccc}
    \toprule
    Method & IoU & MSE & PNSR & LPIPS \\
    \midrule
    RT\_simult & 69.5\%\ & 741.58 & 20.19 & \textbf{0.10}\\
    No Eq.\ref{eq:motion} & 70.1\%\ & 749.58 & 20.14 & \textbf{0.10}\\
    No Eq.\ref{eq:affinity} & 59.4\%\ & 838.11 & 19.47 & 0.11\\
    \midrule
    RAFT corr.& 63.5\%\ & 788.39 & 19.78 & 0.11 \\
    COTR corr.& 60.2\%\ & 862.28 & 19.40 & 0.12 \\
    \midrule
    APES & \textbf{71.0\%} & \textbf{733.12} & \textbf{20.20} & \textbf{0.10}\\
    \bottomrule
  \end{tabular}
   \vspace{-5pt}
  \caption{Ablation study of our variants (OkaySamurai  dataset).}
 \vspace{-12pt}  
  \label{tab:ablation}
\end{table}

We perform a set of ablation experiments on the OkaySamurai dataset since it includes ground-truth articulated parts for evaluation. We compare with the following variants of our method: \textbf{RT\_simult}: we predict rotation and translation of each superpixel simultaneously, instead of sequentially as in our original method. \textbf{No Eq.~\ref{eq:motion}}: we train the segmentation module with Eq.~\ref{eq:affinity} and Eq.~\ref{eq:iouloss} only, without supervision for the motion residual matrix. \textbf{No Eq.~\ref{eq:affinity}}: we train the segmentation module with Eq.~\ref{eq:motion} and Eq.~\ref{eq:iouloss} only, without supervision on the affinity matrix. \textbf{RAFT corr}: Instead of our UNet-based correspondence module, we use RAFT to predict correspondences used in the following steps. \textbf{COTR corr}: we use COTR to produce correspondences instead.

We report all our evaluation metrics in Table~\ref{tab:ablation}, including reconstruction metrics, since all the above variants employ the same part selection and reconstruction stage. We observe inferior results from all reduced variants.

\section{Conclusion}
\label{sec:conclusion}
We presented APES, a method that extracts articulated parts from a sparse set of 
character poses of a sprite sheet. As far as we know, APES is the first method capable of automatically extracting deformable puppets from unsegmented character poses. 
We believe that methods able to parse character artwork and generate rigs have the potential to significantly automate the character animation workflow.

\vspace{-5mm}
\pichskip{4pt}
\parpic[r][b]{
  \begin{minipage}{30mm}
  \vspace{5.5mm}
  \includegraphics[width=1\columnwidth]{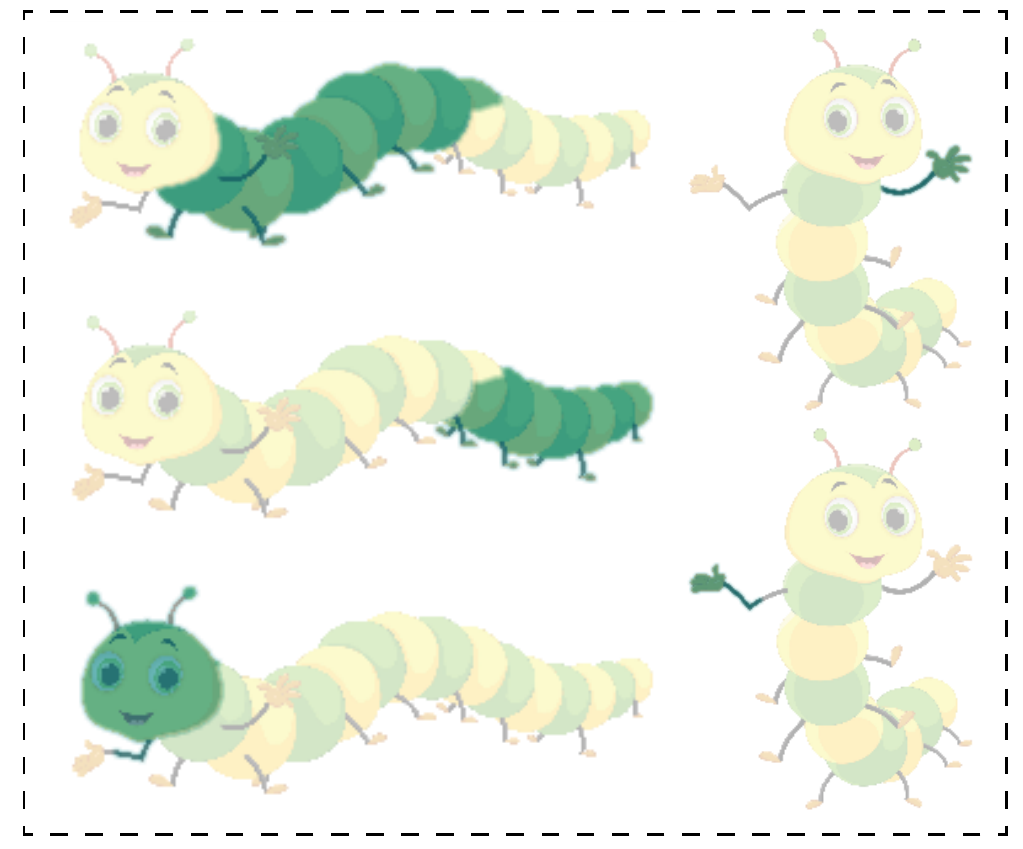}
  \vspace{-7mm}
  \end{minipage}
}
 
\vspace{-1pt}
\paragraph{Limitations.} While we can handle a wide range of character styles and part configurations, there are some limitations.
Parts with non-rigid or subtle motion cannot be extracted well. For example, given the caterpillar poses shown on the right, our method extracts the head, arms, and body chunks, yet it does not segment the thin legs and the individual abdomen segments, since  these do not seem to have distinct rotations with respect to the rest of the body. 
As discussed in our experiments, the boundary of parts is not always accurate. Our method selects parts such that they minimally overlap during part selection. As a result, small articulated parts might be missed and replaced by larger ones.
Handling strongly overlapping parts
more explicitly, layer order changes between sprites (such as a character turning around), and large occlusions
would make our method applicable to a wider range of sprite sheet cases.

\vspace{-7pt}
\paragraph{Acknowledgements.}
Our research was partially funded by NSF (EAGER-1942069) and Adobe.

{\small
\bibliographystyle{ieee_fullname}
\bibliography{egbib}
}

\newpage
\renewcommand\appendixpagename{Appendix}
\appendix
\appendixpage
\section{Applications}
In our supplementary video \url{https://youtu.be/YQtbRXFKNZE}, we show two applications based on extracted articulated parts from APES: part-based animation and puppet creation via part swapping. We refer readers to the video for a demonstration of these applications. 

\begin{figure}[!h]
    \centering
    \begin{subfigure}[t]{0.2\textwidth}
         \centering
         \includegraphics[height=3.25cm]{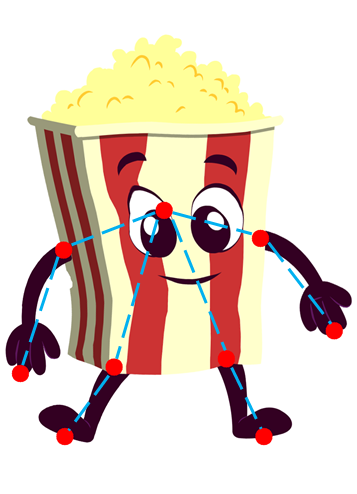}
         \label{fig:skel-1}
     \end{subfigure}
     \begin{subfigure}[t]{0.2\textwidth}
         \centering
         \includegraphics[height=3.25cm]{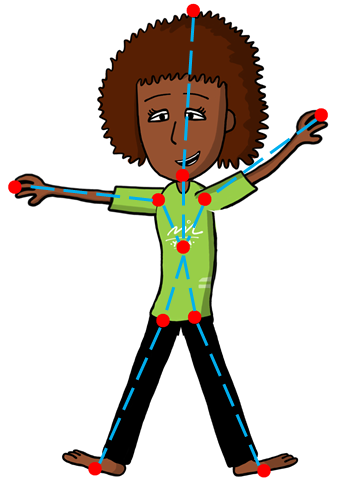}
         \label{fig:skel-2}
     \end{subfigure}
    
    \caption{Control points (red dots) and skeletons created based on the extracted parts from APES.}
    \label{fig:skel}
\end{figure}

For part-based animation, we create a set of control points or joints  based on the extracted parts. We follow a simple heuristic to obtain them. First, we compute the centroid point of the whole character, and designate the part closest to the centroid as the ``central'' part (this coincides with the torso in Fig.~\ref{fig:skel}). We create a joint at the center of this central part. The rest of the parts are designated as ``limbs''.
We create joints (``pin joints'') at the center of the intersection between the limbs and the central part (e.g., hips, shoulders in Fig.~\ref{fig:skel}). Finally, we extract the medial axis of each limb, and create a joint at the medial axis point furthest to the pin joint for each limb (these points correspond to fingers and toes in Fig.~\ref{fig:skel}). The resulting joints form a simple control structure 
(e.g., an animation skeleton) that can be loaded into standard animation software. We use the Adobe's Character Animator in the supplementary video. By manipulating the joint positions and angles, users can animate the characters based on the joints and parts extracted by APES.

\section{Non-rigid Deformation for Reconstruction}
As explained in Section 3.3, our part selection procedure uses a random rounding algorithm to solve the integer linear programming problem. The random rounding algorithm gives us multiple possible solutions. We pick the best solution in terms of reconstruction error after deforming each candidate set of parts to reconstruct the input poses. We describe here the deformation procedure.

\begin{figure}[!h]
    \centering
    \begin{subfigure}[t]{0.115\textwidth}
         \centering
         \includegraphics[width=\textwidth]{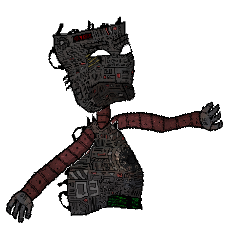}
         \caption{target pose}
         \label{fig:arap-tar}
     \end{subfigure}
     \begin{subfigure}[t]{0.115\textwidth}
         \centering
         \includegraphics[width=\textwidth]{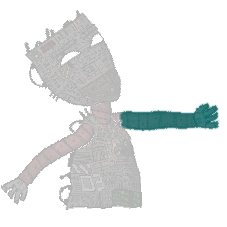}
         \caption{selected part}
         \label{fig:arap-select}
     \end{subfigure}
     \begin{subfigure}[t]{0.115\textwidth}
         \centering
         \includegraphics[width=\textwidth]{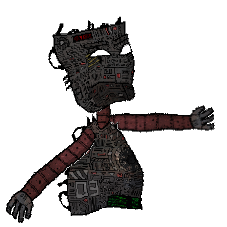}
         \caption{linear trans.}
         \label{fig:arap-linear}
     \end{subfigure}
     \begin{subfigure}[t]{0.115\textwidth}
         \centering
         \includegraphics[width=\textwidth]{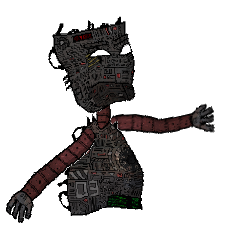}
         \caption{ARAP-based}
         \label{fig:arap-recon}
     \end{subfigure}
    
    \caption{Example of reconstruction via non-rigid deformation. 
    By using the best-fit rigid transformation, the deformation (c) of the left arm  cannot reconstruct well the target pose (a). Using the  ARAP-based reconstruction (d), the arm is aligned better to the target appearance.}
    \label{fig:arap}
\end{figure}

Let $\bQ = \{\bq_c\}_{c=1}^C$ be a set of selected parts (where $C$ is their total number), and the image $\bI_t$ be the target pose to be reconstructed. A straight-forward method for reconstruction is fitting the optimal translation and rotation transformation for each part based on the predicted correspondences using the Procrustes orthogonal analysis used in ICP methods \cite{besl1992method}. 
 Such method is computationally efficient, but is sensitive to the noise in the predicted correspondences, and cannot capture non-rigid deformations that may exist in the input poses.
 Instead, we follow an as-rigid-as-possible deformation procedure (ARAP~\cite{sorkine2007rigid}). The ARAP takes as input a control mesh, and target positions for one or more of its vertices. To create a control mesh for each part $\bq_c \in \bQ$, we first uniformly sample a set of vertices $\V_c$ from its oriented bounding box. Then we create the Delaunay triangulation of the vertices. Each control mesh also incorporates the texture from the original appearance of its associated part. We treat the target positions $\{\V'_c\}_{c=1}^C$ of the mesh vertices as unknowns in an optimization problem that attempts to deform the control meshes of all parts such that their resulting appearance is as close as possible to the target pose $\bI_t$, while at the same time the part deformations are as-rigid-as-possible. Specifically, we solve the following problem:
\vspace{-5mm}
\begin{equation}
    L(\V'_c) =\|\sum\limits_{c=1}^C
    \Phi(\V'_c)-\bI_t\|^2_2 +
    \lambda_r 
    \sum\limits_{c=1}^C \R(\V_c, \V'_c)
\label{eq:arap}
\end{equation}
where $\lambda_r$ is set to $0.05$ in our experiments. The first term measures the reconstruction error between the deformed parts (caused by the shifted vertices) and the target pose. $\Phi$ renders each part with the shifted control vertices 
 $\bV'_c$ based on barycentric coordinates and bilinear interpolation~\cite{lu2020}, which is differentiable. The second term 
is a regularization term that preserves the local shape during deformation. Specifically, it penalizes deviation from local rigid deformation for each control vertex neighborhood
~\cite{sorkine2007rigid}:
\begin{equation}
    \R(\V_c, \V'_c) = 
    \sum_{\bv_i \in \V_c}
    \sum_{\bv_j \in \mathcal{N}(\bv_i)}\|    (\bv'_j-\bv'_i) - \bR_i(\bv_j-\bv_i)\|^2
\end{equation}
where $\mathcal{N}(\bv_i)$ is the neighborhood of the control vertex $\bv_i$ and $\bR_i$ is the best fit rotation of its neighborhood to the deformed configuration. The best fit rotations are computed via orthogonal Procrustes analysis. The optimization problem is solved iteratively. At each iteration, we alternative between solving for the deformed control vertices, and optimal rotations, as proposed in \cite{sorkine2007rigid}. Fig.\ref{fig:arap} shows an example of our reconstruction approach.

\paragraph{Non-rigid deformation for augmentation.} As discussed in Section 5.1, we apply small, non-rigid
deformations on each training body part to improve the pose
diversity during training for the OkaySamurai dataset. To do so,  for each part we uniformly sample control vertices in its oriented bounding box, and randomly shift the points by offsets sampled from Gaussian distribution $\mathcal{N}(0, \sigma^2)$ where $\sigma$ is set as 2\% of the maximum image dimension. We form the  control meshes for parts, and deform them to reach the shifted control vertices using an ARAP deformation similar to Eq.~\ref{eq:arap}.; here the reconstruction error measures difference between deformed and original control vertex positions.

\section{Architecture Details}
We provide here additional details of our network architecture.

\paragraph{Correspondence Module.}
Table~\ref{table:corr_arch} lists the layers used in the correspondence module along with the size of their output map. We call the architecture of our module as  ``Gated UNet'' since the convolutional layers in the encoder implement Gated Convolution~\cite{yu2018free}.

\begin{table}[t]
\centering 
\caption{\textbf{Correspondence module architecture (Gated UNet).} \textit{Conv3x3} is convolutional layer with kernel size 3. \textit{LN} is LeakyReLU with negative slope as 0.2. \textit{BN} is BatchNorm. \textit{Dot} between LR and Sigmoid is element-wise product. \textit{Upsample} uses bilinear interpolation for upsampling. We note that there are skip connections between corresponding layers from encoder and decoder, following the original U-Net architecture.}
\footnotesize
\begin{tabular}{|@{}c@{}|@{}c@{}|@{}c@{}|}
\hline

    & Layers                           & Output   \\
\hline
\hline
Input    &  Concat(image, mask)    &256$\times$256$\times$4  \\\cline{2-3}
    
\hline
\hline
\multirow{14}{*}{Encoder} & BN(LR(Conv3x3)$\cdot$Sigmoid(Conv3x3))
                          & 256$\times$256$\times$32 \\\cline{2-3}
                          & BN(LR(Conv3x3)$\cdot$Sigmoid(Conv3x3))
                          & 256$\times$256$\times$32 \\\cline{2-3}
                          & MaxPooling(2)
                          & 128$\times$128$\times$32 \\\cline{2-3}
                          
                          & BN(LR(Conv3x3)$\cdot$Sigmoid(Conv3x3))              &128$\times$128$\times$64 \\\cline{2-3}
                          & BN(LR(Conv3x3)$\cdot$Sigmoid(Conv3x3))              &128$\times$128$\times$64 \\\cline{2-3}
                          & MaxPooling(2)
                          & 64$\times$64$\times$64 \\\cline{2-3}
                          
                          & BN(LR(Conv3x3)$\cdot$Sigmoid(Conv3x3))              &64$\times$64$\times$128 \\\cline{2-3}
                          & BN(LR(Conv3x3)$\cdot$Sigmoid(Conv3x3))              &64$\times$64$\times$128 \\\cline{2-3}
                          & MaxPooling(2)
                          & 32$\times$32$\times$128 \\\cline{2-3}
                          
                          & BN(LR(Conv3x3)$\cdot$Sigmoid(Conv3x3))              &32$\times$32$\times$256 \\\cline{2-3}
                          & BN(LR(Conv3x3)$\cdot$Sigmoid(Conv3x3))              &32$\times$32$\times$256 \\\cline{2-3}
                          & MaxPooling(2)
                          & 16$\times$16$\times$256 \\\cline{2-3}
                          
                          & BN(LR(Conv3x3)$\cdot$Sigmoid(Conv3x3))              &16$\times$16$\times$256\\\cline{2-3}
                          & BN(LR(Conv3x3)$\cdot$Sigmoid(Conv3x3))              &16$\times$16$\times$256 \\
\hline
\hline
\multirow{13}{*}{Decoder} & Upsample(2)
                          & 32$\times$32$\times$256 \\\cline{2-3}
                          & BN(ReLU(Conv3x3))
                          & 32$\times$32$\times$128 \\\cline{2-3}
                          & BN(ReLU(Conv3x3))
                          & 32$\times$32$\times$128 \\\cline{2-3}
                          
                          & Upsample(2)
                          & 64$\times$64$\times$128 \\\cline{2-3}
                          & BN(ReLU(Conv3x3))               &64$\times$64$\times$64 \\\cline{2-3}
                          & BN(ReLU(Conv3x3))              &64$\times$64$\times$64 \\\cline{2-3}
                          
                          & Upsample(2)
                          & 128$\times$128$\times$64 \\\cline{2-3}
                          & BN(ReLU(Conv3x3))              &128$\times$128$\times$32 \\\cline{2-3}
                          & BN(ReLU(Conv3x3))              &128$\times$128$\times$32 \\\cline{2-3}
                          
                          & Upsample(2)
                          & 256$\times$256$\times$32 \\\cline{2-3}
                          & BN(ReLU(Conv3x3))              &256$\times$256$\times$32 \\\cline{2-3}
                          & BN(ReLU(Conv3x3))              &256$\times$256$\times$32 \\\cline{2-3}
                          
                          & BN(ReLU(Conv3x3))              &256$\times$256$\times$64\\

\hline 
\end{tabular}
\label{table:corr_arch}
\end{table}

\paragraph{Clustering Module.}
Table~\ref{table:cluster_arch} shows the architecture of the clustering module.

\begin{table}[t]
\centering
\caption{\textbf{Clustering module architecture.} The symbol ``\textit{pred.R}'' means predicted rotations, and ``\textit{pred.T}'' means predicted translations. To update the voting map, we apply the predicted rotations to source pixels.}
\footnotesize
\begin{tabular}{|c|c|c|}
\hline

    & Layers                           & Output   \\
\hline
\hline
Input    &  Concat(voting map, mask)    &256$\times$256$\times$5  \\\cline{2-3}

\hline
\hline
Gated UNet             &  \makecell{similar to Table \ref{table:corr_arch} with \\ intermediate channel numbers \\as 16, 32, 64, 128, 256}
                          & 256$\times$256$\times$16 \\
\hline
\makecell{Average pooling \\ per superpixel}
                          &  N/A
                          & $K_s\times$16 \\ 

\hline
MLP             &  16$\rightarrow$64$\rightarrow$2
                          & pred.R: $K_s\times$2 \\

\hline
\hline
Updated input    &  Concat(updated voting map, mask)    &256$\times$256$\times$5  \\\cline{2-3}      

\hline
\hline
Gated UNet             &  \makecell{similar to Table \ref{table:corr_arch} with \\ intermediate channel numbers \\as 16, 32, 64, 128, 256}
                          & 256$\times$256$\times$16 \\
\hline
\makecell{Average pooling \\ per superpixel}
                          &  N/A
                          & $K_s\times$16 \\ 

\hline
MLP             &  16$\rightarrow$64$\rightarrow$2
                          & pred.T: $K_s\times$2 \\

\hline 
\end{tabular}
\label{table:cluster_arch}
\end{table}

\section{Implementation Details}
The correspondence module is first trained alone. We set the learning rate to $10^{-3}$ and decrease it to $10^{-4}$ after $5$ epochs. Then we train both the correspondence and clustering modules using the soft nearest neighbor in Equation 5 of the main text. We set the learning rate to $10^{-6}$ for the correspondence module and $10^{-4}$ for the segmentation module with a batch size of $8$ for this stage. We use the Adam optimizer. The height $H$ and width $W$ of the images and voting maps are always $256$. The number of clusters $C_s$ during training is set to $12$. 

\section{Correspondence Comparison}
\begin{figure}[!h]
    \centering
    \includegraphics[width=1.0\linewidth]{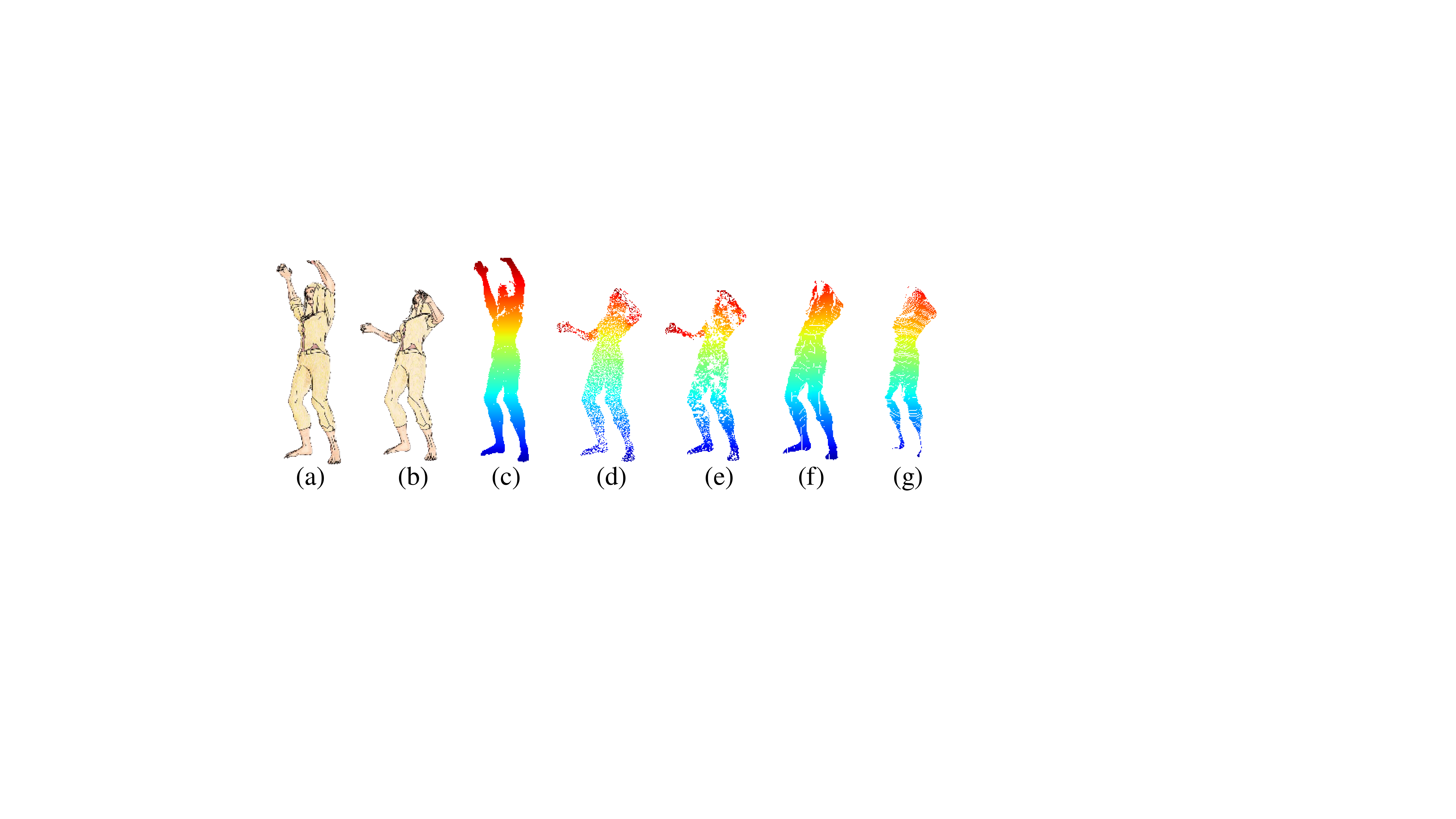}
    \caption{Visualization of the predicted correspondences from our method, RAFT and COTR. From left to right: (a) source image (b) target image  (c) color-coded source pixels (d) ground-truth target pixels (e) prediction of our method (f) prediction of RAFT (g) prediction of COTR. Corresponding pixels have same color in the visualization of correspondence maps. Our method matches pixels more accurately compared to other methods.}
    \label{fig:comp_cf}
\end{figure}

Fig.~\ref{fig:comp_cf} shows a qualitative comparison example of predicted correspondences between our method, RAFT \cite{teed2020raft} and COTR \cite{jiang2021cotr}. To help RAFT and COTR better use the foreground masks, we map the predicted target positions to their nearest neighbors in the foreground. RAFT and COTR cannot produce correspondences reliably e.g., for hand and head regions, as shown below. 

\section{More Qualitative Results}

\begin{figure}[!h]
  \centering
  \includegraphics[width=1.0\linewidth]{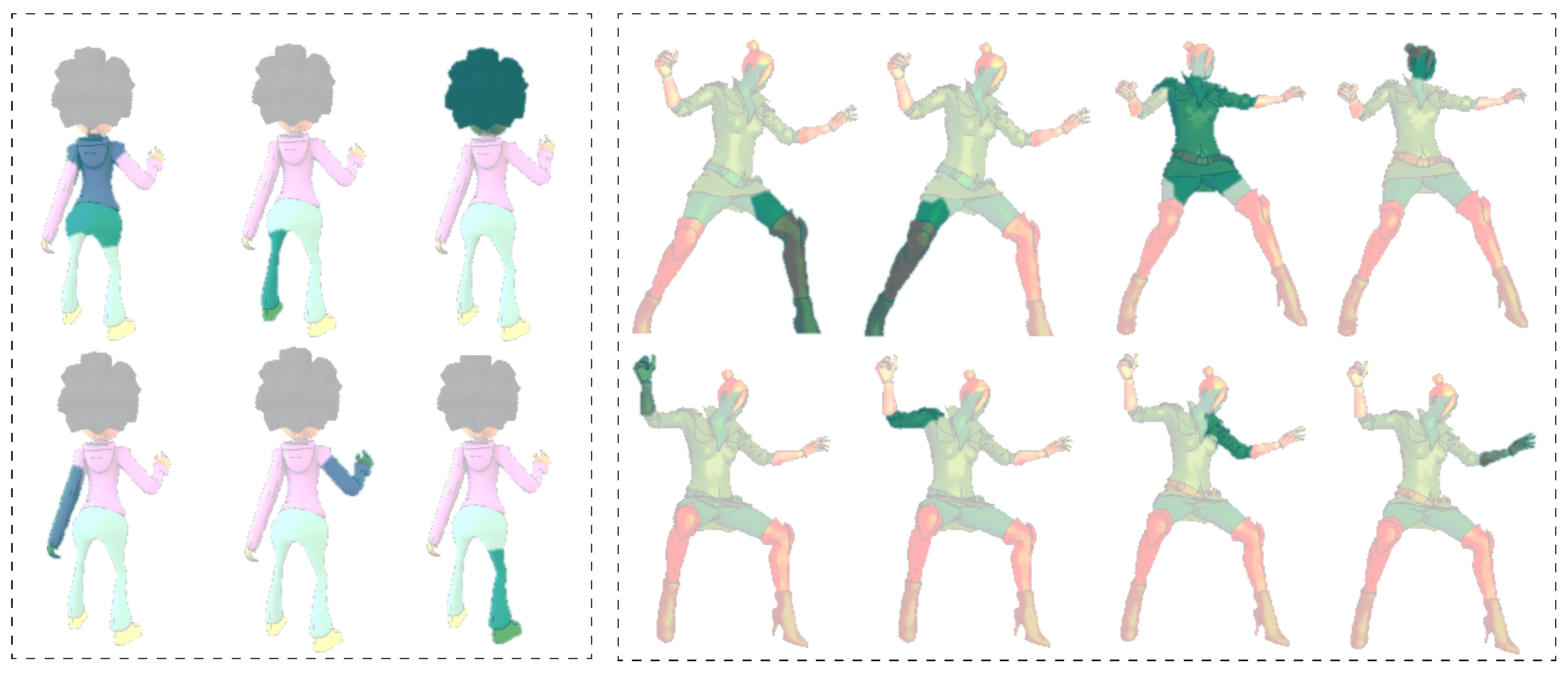}
  \caption{Part extraction results in the CreativeFlow+ dataset.}
  \label{fig:cf_seg}
\end{figure}

 We show here additional qualitative results from the The CreativeFlow+ dataset. We note that this 
dataset does not include segmentations of articulated parts. Their provided segmentation maps are based on mesh components, which often do not match articulation. Thus, as an additional test set, we  manually segmented $12$ characters into rigid parts from their test split,
and used them as reference
to measure the performance of our trained model on them. We achieve an IoU of $67\%$, which is slightly lower than the IoU we achieved for the OkaySamurai dataset ($71\%$). Two examples are shown in Fig.~\ref{fig:cf_seg}.

We include the results for the test sprite sheets in OkaySamurai dataset and SPRITE in our code repository.

\end{document}